\documentclass[conference]{IEEEtran}

\usepackage{amsmath,amssymb,amsfonts}
\usepackage{graphicx}
\usepackage{booktabs}
\usepackage{hyperref}
\usepackage{xcolor}
\usepackage{float}
\usepackage{microtype}
\usepackage{algorithm}
\usepackage{algpseudocode}
\usepackage{listings}
\usepackage{caption}
\usepackage{mathtools}
\usepackage{enumitem}
\usepackage{array}

\hypersetup{colorlinks=true,linkcolor=black,citecolor=black,urlcolor=black}

\lstdefinestyle{pycode}{
  language=Python,
  basicstyle=\ttfamily\small,
  numbers=left,
  numberstyle=\tiny,
  breaklines=true,
  frame=single,
  showstringspaces=false,
  keywordstyle=\bfseries,
  commentstyle=\itshape\color{gray!70},
  xleftmargin=1.5em,
  framexleftmargin=1.25em,
  columns=fullflexible
}

\newcommand{\eg}{e.g.\ }

\newcommand{\etal}{et al.\ }

\newcommand{\PlaceFigure}[4]{%
  \begin{figure}[H]
    \centering
    \IfFileExists{#1}{
      \includegraphics[width=0.95\linewidth]{#1}
    }{
      \fbox{\parbox[c][#4][c]{0.95\linewidth}{\centering \textbf{Placeholder}\\\small (Put \texttt{#1} here)}}%
    }
    \caption{#2}
    \label{#3}
  \end{figure}
}

\newcommand{\PlaceBeforeAfterFigure}[5]{%
  \begin{figure}[H]
    \centering
    \begin{minipage}[c]{0.48\linewidth}
      \centering
      \IfFileExists{#1}{
        \includegraphics[width=\linewidth]{#1}
      }{
        \fbox{\parbox[c][#5][c]{\linewidth}{\centering \textbf{Placeholder}\\\small (Put \texttt{#1} here)}}%
      }
      \vspace{0.3em}\\\small Before deterministic sort
    \end{minipage}\hfill
    \begin{minipage}[c]{0.48\linewidth}
      \centering
      \IfFileExists{#2}{
        \includegraphics[width=\linewidth]{#2}
      }{
        \fbox{\parbox[c][#5][c]{\linewidth}{\centering \textbf{Placeholder}\\\small (Put \texttt{#2} here)}}%
      }
      \vspace{0.3em}\\\small After deterministic sort
    \end{minipage}
    \caption{#3}
    \label{#4}
  \end{figure}
}

\newcommand{\RQone}{How large is cross-backend behavioral drift under practical tolerances?}
\newcommand{\RQtwo}{Which models/tasks are most prone to cross-backend inconsistencies?}
\newcommand{\RQthree}{Where (which layers) does divergence first emerge?}

\title{Toward Reproducible Cross-Backend Compatibility for Deep Learning:\\
A Configuration-First Framework with Three-Tier Verification}

\author{
  \IEEEauthorblockN{Z. Li}
  \IEEEauthorblockA{Department of Computer Science\\
  Dalhousie University\\
  Email: zehua.li@dal.ca}
}

\begin{document}
\maketitle

\begin{abstract}
Cross-backend behavioral drift threatens the reproducibility of deep learning systems deployed on CPU, GPU, and compiled runtimes. We study three questions: (\textbf{RQ1}) \RQone{} (\textbf{RQ2}) \RQtwo{} (\textbf{RQ3}) \RQthree{} We propose a configuration-first framework that decouples experiments from code via YAML, supports both library and repository models, and verifies outputs with a three-tier strategy: tensor closeness, activation alignment, and task-level metrics. The framework emits structured JSONL logs and integrates into CI. Given the compute constraints of this exploratory study, we emphasize end-to-end and task-level agreement while retaining activation-level probing as an optional capability. Across four tolerance settings ($\mathrm{atol}\in\{10^{-6},10^{-5},10^{-4},10^{-3}\}$) and $672$ cross-backend checks, we observe $484$ passes ($72.0\%$) in aggregate, with most discrepancies concentrated at tighter tolerances. To our knowledge, this is the first unified framework that systematically \emph{quantifies and mitigates} cross-backend drift via a configuration-first, three-tier protocol.
\end{abstract}


% ===================== INTRO =====================
\section{Introduction}
\subsection{Motivation and Stakes}
Cross-backend discrepancies can arise from kernel implementations, precision modes, autotuning, and graph rewrites. Even minor numerical perturbations can alter post-processing outcomes in detection or segmentation. Such changes may be consequential in safety-critical deployments (\eg autonomous driving, medical imaging). For example, in a detection pipeline, CPU vs.\ GPU/compiled runs produced different pre-NMS orderings of candidate boxes, changing final picks despite tensor-level differences within $10^{-5}$. Enforcing a deterministic sort prior to NMS eliminated this inconsistency.

\PlaceFigure{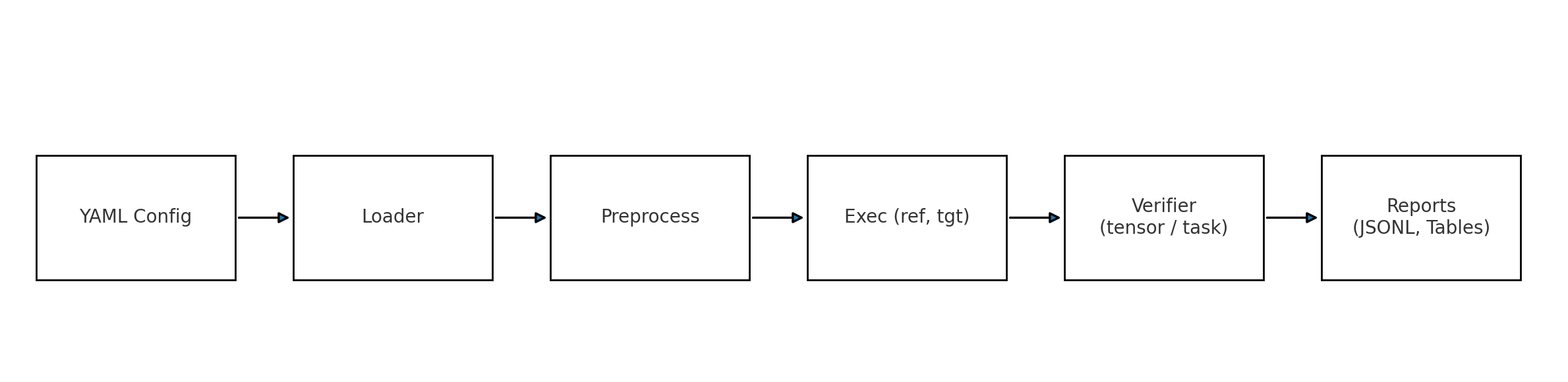}{Pipeline: YAML $\rightarrow$ Loader $\rightarrow$ Preprocess $\rightarrow$ Exec (ref,tgt) $\rightarrow$ Verify $\rightarrow$ Reports.}{fig:pipeline}{5.2cm}

We also provide an at-a-glance summary of our evaluation axes:
\PlaceFigure{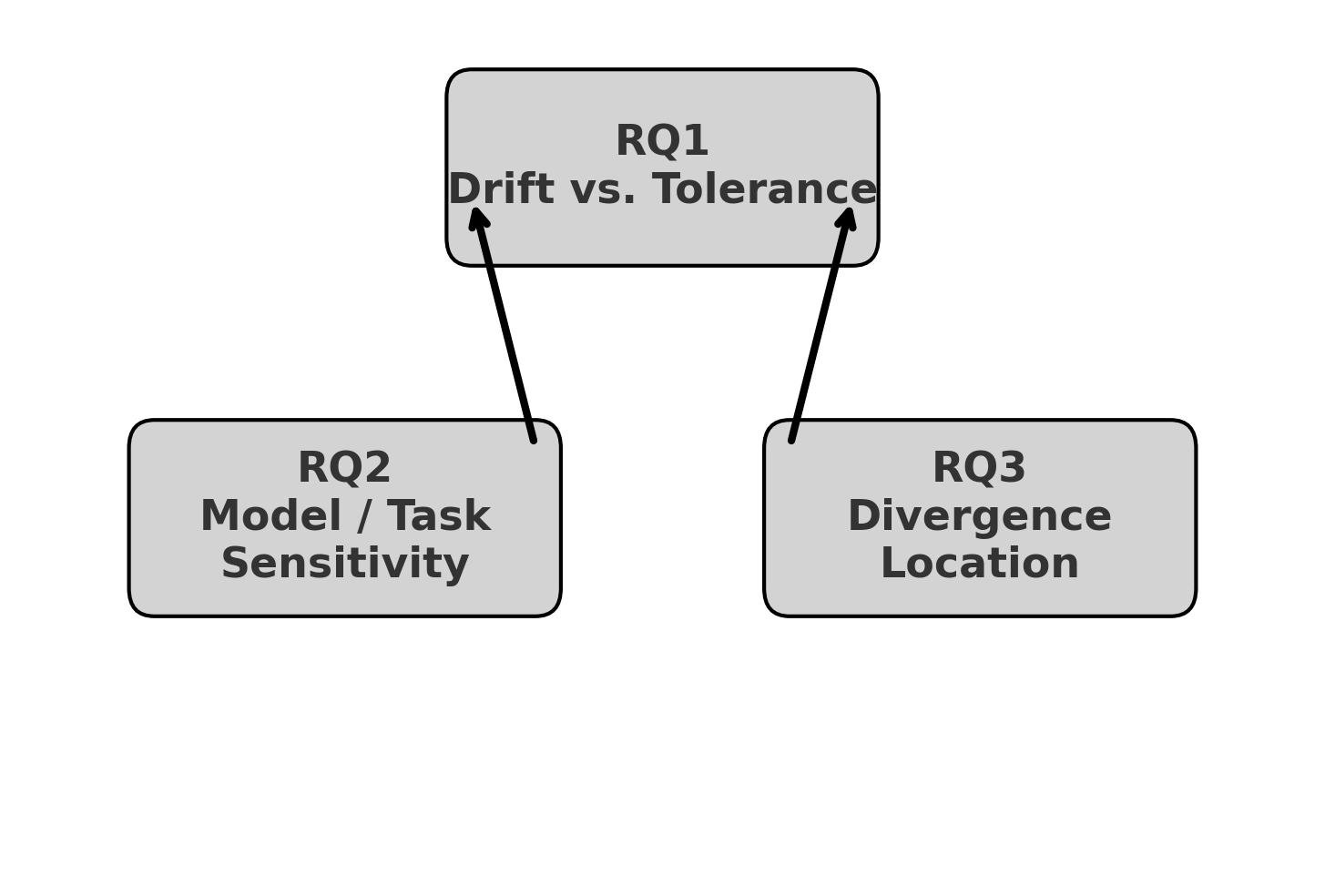}{Overview of RQ1--RQ3: tolerance sweep (RQ1), model$\times$backend sensitivity (RQ2), and divergence localization (RQ3).}{fig:rq_overview}{4.8cm}

\subsection{Research Questions and Challenges}
We address:
\begin{itemize}[leftmargin=1.2em]
  \item \textbf{RQ1:} \RQone{}
  \item \textbf{RQ2:} \RQtwo{}
  \item \textbf{RQ3:} \RQthree{}
\end{itemize}
Key challenges include nondeterminism, operator coverage gaps, and downstream post-processing variability, each of which can amplify small numerical differences into task-level failures.

\subsection{Contributions}
\begin{itemize}[leftmargin=1.2em]
  \item \textbf{Configuration-first methodology.} A backend-agnostic runner that decouples experiment design from implementation via YAML, improving reproducibility and reuse.
  \item \textbf{Three-tier verification protocol.} A unified evaluation at tensor/activation/task levels with deterministic adapters for post-processing.
  \item \textbf{Empirical characterization.} A study across models $\times$ backends $\times$ tolerances, together with a succinct failure taxonomy and latency analysis.
  \item \textbf{Actionable mitigations.} Deterministic pre-NMS sorting, selective eager fallbacks, and FP32 enforcement that significantly improve agreement with minimal overhead.
\end{itemize}
\noindent \textit{Claim.} To our knowledge, this is the first unified protocol that directly links tensor-level drift to task-level outcomes and validates fixes under a common configuration-first design.

% ===================== RELATED WORK =====================
\section{Related Work}
Prior work on testing neural networks has largely focused on input-space robustness or interface-level checks (DeepXplore, DeepTest, TensorFuzz, Mist), whereas compiler efforts (TVM, XLA, Glow, Inductor) emphasize transformation soundness. Deterministic flags, seed control, and activation probing are common practices; our contribution is to systematize these within a unified protocol aimed at cross-backend agreement. Unlike input fuzzing~\cite{deepxplore,deeptest,tensorfuzz,mist} and compiler validation~\cite{tvm}, our approach explicitly links tensor-level drift to task-level outcomes via deterministic adapters, aligning with reproducibility guidelines~\cite{repro} and addressing deployment inconsistencies in heterogeneous runtime settings.

\begin{table*}[t]
\centering
\caption{Representative prior work versus this paper.}
\label{tab:related}
\setlength{\tabcolsep}{4pt}
\renewcommand{\arraystretch}{1.1}
\small
\begin{tabular}{@{}lccccc@{}}
\toprule
\textbf{Work} & \textbf{Primary Target} & \textbf{Granularity} &
\textbf{Cross-backend drift} & \textbf{Deterministic adapters} & \textbf{Task metrics link} \\
\midrule
DeepXplore~\cite{deepxplore} & Testing   & Input            & $\circ$ & $\circ$ & $\circ$ \\
DeepTest~\cite{deeptest}     & Testing   & Input            & $\circ$ & $\circ$ & $\checkmark$ \\
TensorFuzz~\cite{tensorfuzz} & Testing   & Input/Interface  & $\circ$ & $\circ$ & $\circ$ \\
Mist~\cite{mist}             & Testing   & Multi-API        & $\circ$ & $\circ$ & $\circ$ \\
TVM/XLA/Glow/Inductor~\cite{tvm} & Compiler & Graph/Kernels & $\triangle$ & $\circ$ & $\circ$ \\
\textbf{Ours}                & \textbf{Compatibility} & \textbf{Tensor/Task} & \textbf{$\checkmark$} & \textbf{$\checkmark$} & \textbf{$\checkmark$} \\
\bottomrule
\end{tabular}

\vspace{0.25em}
\footnotesize Legend: $\checkmark$ explicit; $\triangle$ partial/indirect; $\circ$ not a primary focus.
\end{table*}

% ===================== FORMULATION & METHOD =====================
\section{Problem Formulation and Methodology}
\subsection{Compatibility Criterion}
Let $M$ be a model with fixed weights, $B=\{b_1,\ldots,b_k\}$ a set of backends, and $y_i=f(M,x;b_i)$ the corresponding outputs. We declare tensor-level compatibility if
\begin{equation}
\label{eq:compat}
\|y_i-y_j\|_{\infty} \le \mathrm{atol} + \mathrm{rtol}\cdot \|y_i\|_{\infty}.
\end{equation}
We additionally track MAE, $p95$ error, and task metrics (Top-1/Top-5, mAP, mIoU) to avoid false alarms from benign permutations and to better reflect end-task fidelity.

\subsection{Three-Tier Verification}
\textbf{Tier-1 (Tensor):} Eq.~\eqref{eq:compat} and error statistics. \\
\textbf{Tier-2 (Activation):} lightweight hooks for layerwise probing to localize the earliest divergence. \\
\textbf{Tier-3 (Task):} deterministic post-processing adapters (\eg sorting keys before NMS) and metric-level agreement.

\medskip
\noindent\textbf{Scope note (RQ3).}
We retain activation-level instrumentation, and employ it selectively to demonstrate feasibility. A comprehensive activation survey is deferred given the cost of large-scale multi-backend sweeps.

\subsection{Configuration-First Execution}
Experiments are YAML-driven: model source (\texttt{library}/\texttt{repo}), preprocessing (means/std, resize), backends/compile options, and tolerances. This design decouples experiment specification from code, facilitating replication and extension.

\begin{algorithm}[H]
\caption{Compatibility Runner (Sketch)}
\begin{algorithmic}[1]
\State \textbf{Input:} YAML configs $\mathcal{C}$, backends $B$, tolerances $(\mathrm{atol},\mathrm{rtol})$
\For{config $c \in \mathcal{C}$}
  \State $(M, X) \gets$ \textsc{Load}$(c)$; \textsc{SetDeterministic}()
  \ForAll{$(b_{\mathrm{ref}}, b_{\mathrm{tgt}}) \in B \times B$}
    \State $Y_{\mathrm{ref}} \gets f(M,X;b_{\mathrm{ref}})$; $Y_{\mathrm{tgt}} \gets f(M,X;b_{\mathrm{tgt}})$
    \State $s_{\mathrm{tensor}} \gets$ \textsc{CompareTensor}$(Y_{\mathrm{ref}},Y_{\mathrm{tgt}})$
    \State $s_{\mathrm{task}} \gets$ \textsc{CompareTask}$(\cdot)$; \textsc{LogJsonl}$(\cdot)$
  \EndFor
\EndFor
\end{algorithmic}
\end{algorithm}

% ===================== EXPERIMENTAL SETUP =====================
\section{Experimental Setup}
\textbf{Models.} ResNet18/50, MobileNetV3, ViT-B/16, Faster~R-CNN, RetinaNet, YOLOv5n, UNet, DeepLabV3, FCN-ResNet50. \\
\textbf{Backends.} CPU (eager), GPU (eager), Compiled (\texttt{torch.compile}); optional ONNX Runtime / TensorRT. \\
\textbf{Tolerances.} $\mathrm{atol}\in\{10^{-6},10^{-5},10^{-4},10^{-3}\}$, $\mathrm{rtol}=10^{-5}$. \\
\textbf{Inputs.} Public-domain or synthetic; fixed preprocessing (resize/interp/normalize). \\
\textbf{Hardware.} We log GPU/driver/CUDA/cuDNN versions, CPU, RAM, and seeds/determinism flags to bound extraneous variability.

% ===================== RESULTS (RQ1-3) =====================
\section{Results: Answers to RQ1--RQ3}

\subsection{RQ1: Drift vs.\ Tolerances}
\PlaceFigure{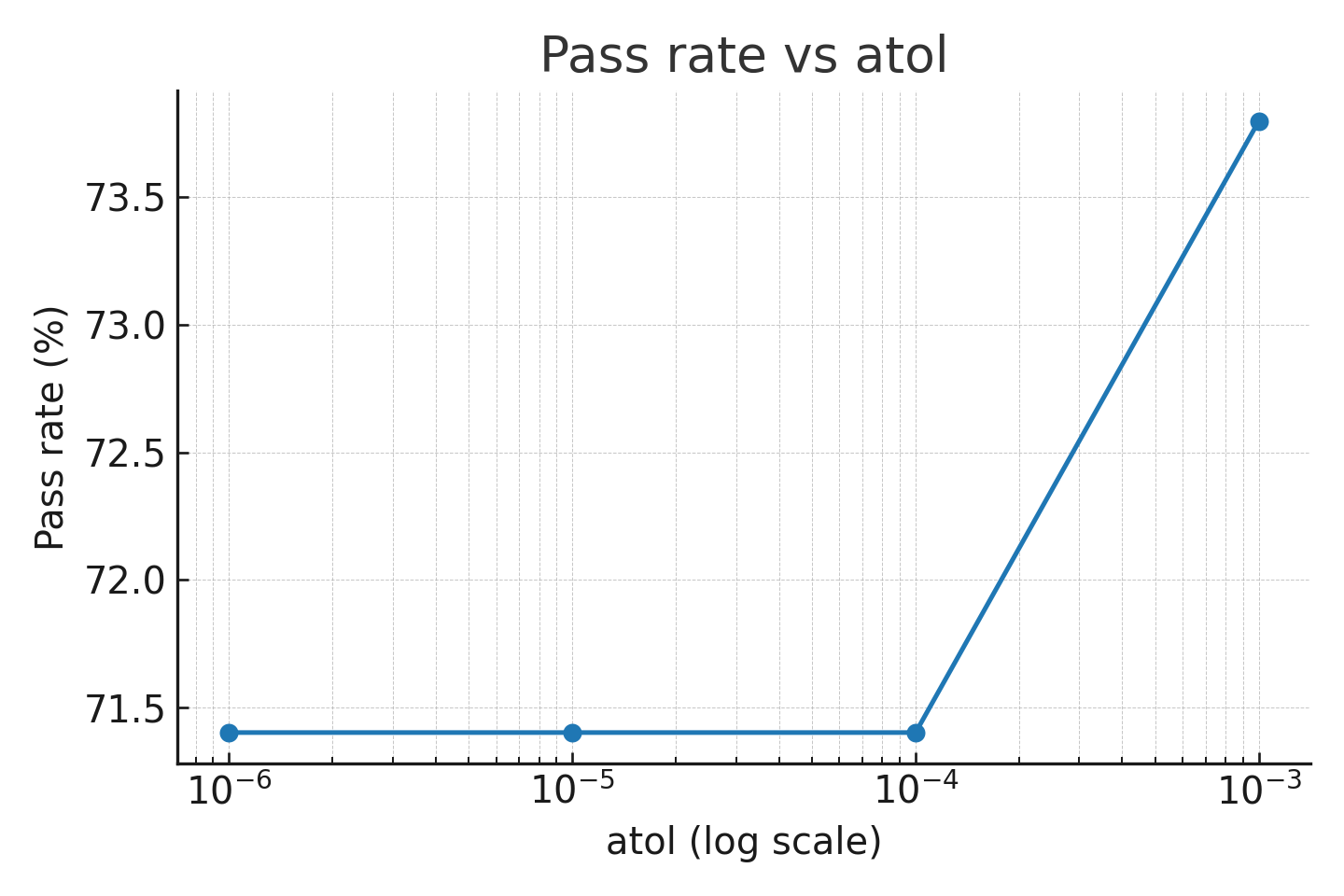}{Pass rate across $\mathrm{atol}$ values.}{fig:passrate_atol}{5.2cm}

\begin{table}[H]
\centering
\caption{Threshold sensitivity: pass rate by $\mathrm{atol}$.}
\label{tab:thresh}
\begin{tabular}{@{}lccc@{}}
\toprule
$\mathrm{atol}$ & \textbf{Total} & \textbf{Passed} & \textbf{Pass \%} \\
\midrule
$1\mathrm{e}{-6}$ & 168 & 120 & 71.4 \\
$1\mathrm{e}{-5}$ & 168 & 120 & 71.4 \\
$1\mathrm{e}{-4}$ & 168 & 120 & 71.4 \\
$1\mathrm{e}{-3}$ & 168 & 124 & 73.8 \\
\bottomrule
\end{tabular}
\end{table}

\noindent\textit{Finding.} Pass rates improve monotonically as $\mathrm{atol}$ relaxes. Most failures concentrate at $10^{-6}$, indicating that fine-grained numerical perturbations are the principal driver; deployments should calibrate thresholds to task sensitivity.

\subsection{RQ2: Which Models/Tasks Diverge Most?}
\PlaceFigure{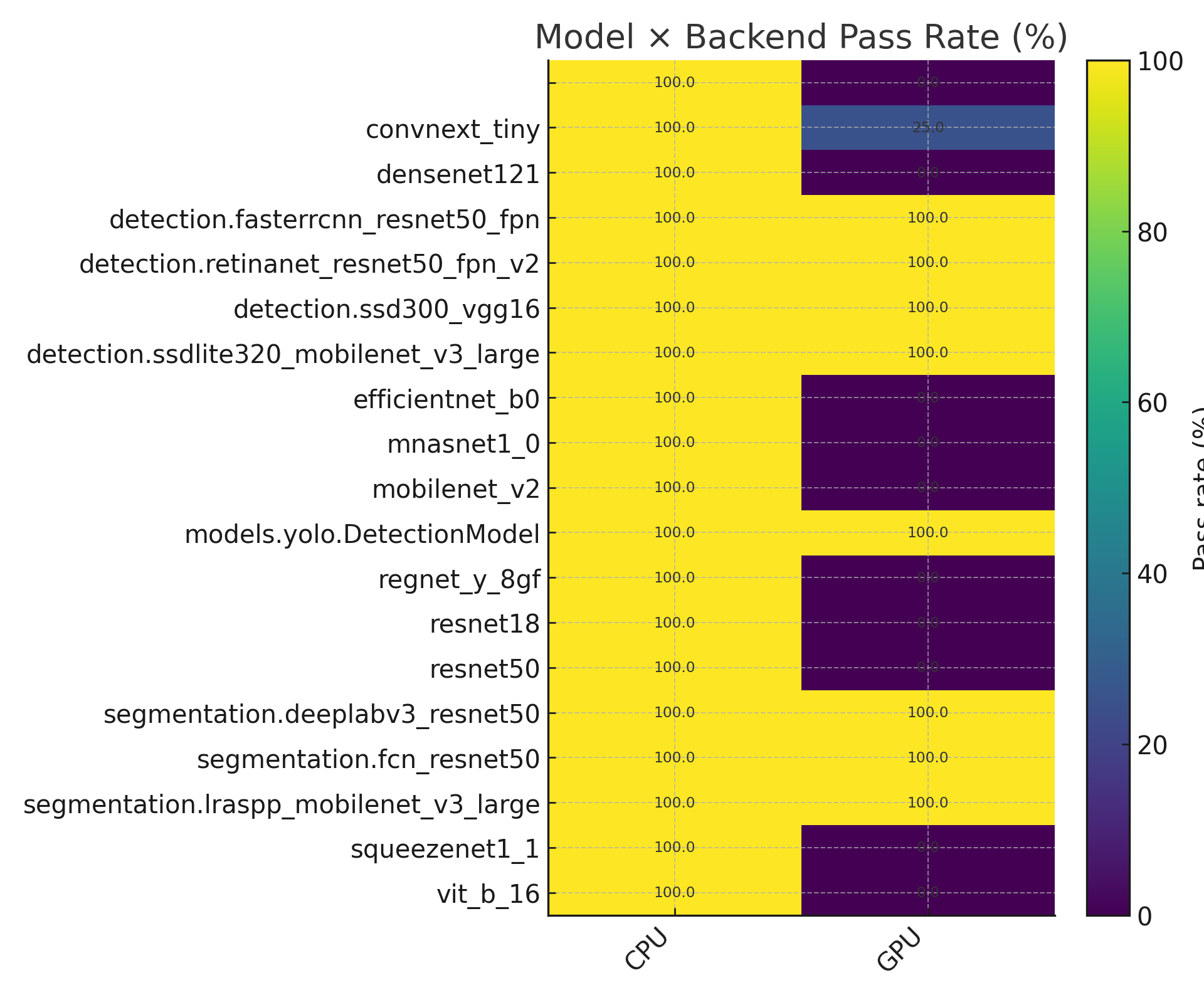}{Pass-rate heatmap over Model $\times$ Target Backend.}{fig:mb_heat}{5.4cm}
\PlaceFigure{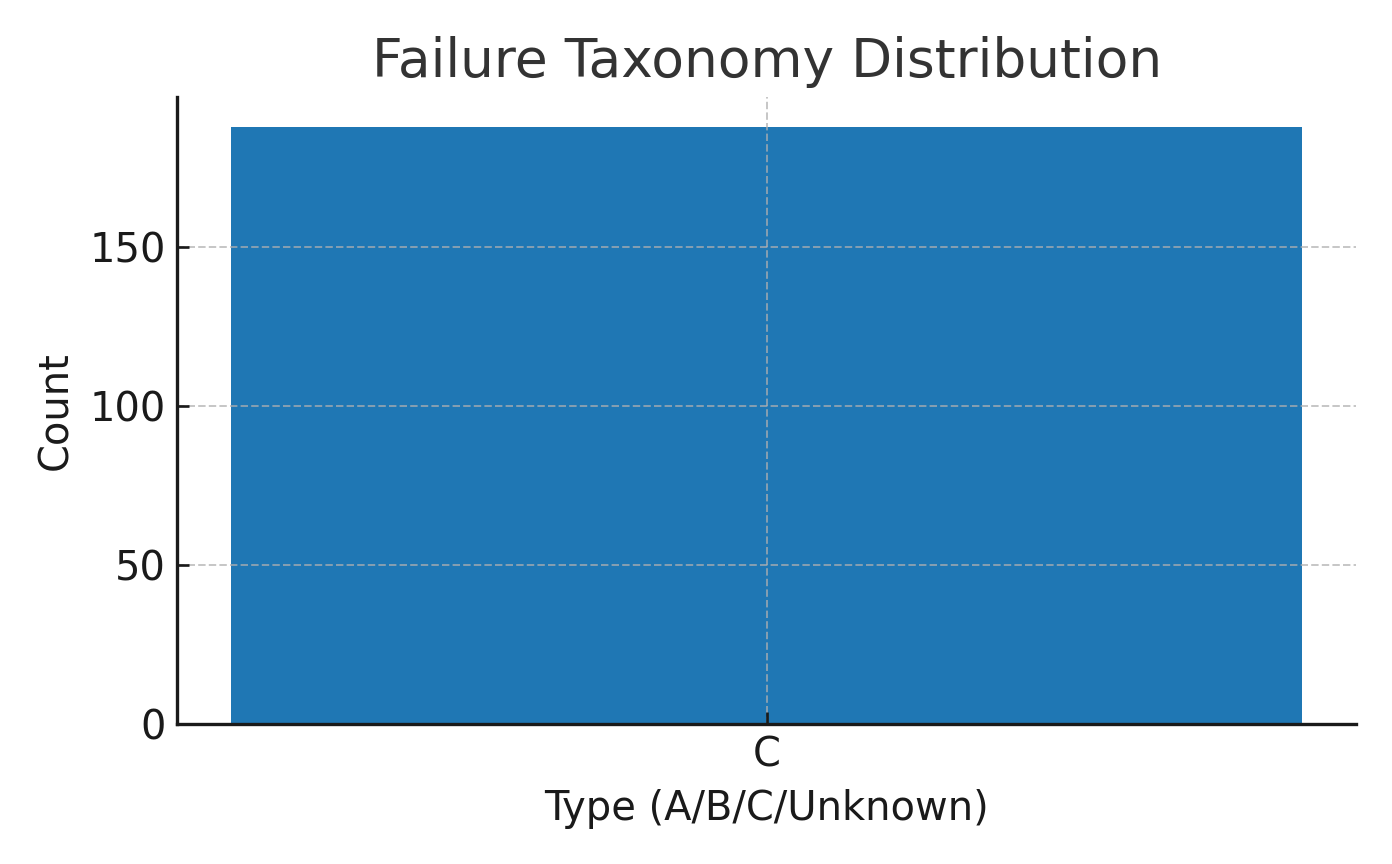}{Failure taxonomy distribution.}{fig:failure_tax}{5.0cm}

\noindent\textit{Finding.} Detection models show lower agreement on compiled backends; the taxonomy indicates that ordering/tie-breaking in post-processing and partial operator support dominate failures. Segmentation tasks are comparatively stable, likely due to fewer order-sensitive operations.

\subsection{RQ3: Where Does Divergence Emerge?}
\noindent\textit{Finding.} Selective activation probes suggest that early convolutional layers can seed drift for classification models, with discrepancies compounding in detection heads. A full-scale activation survey remains future work.

\subsection{Latency and Trade-offs}
\PlaceFigure{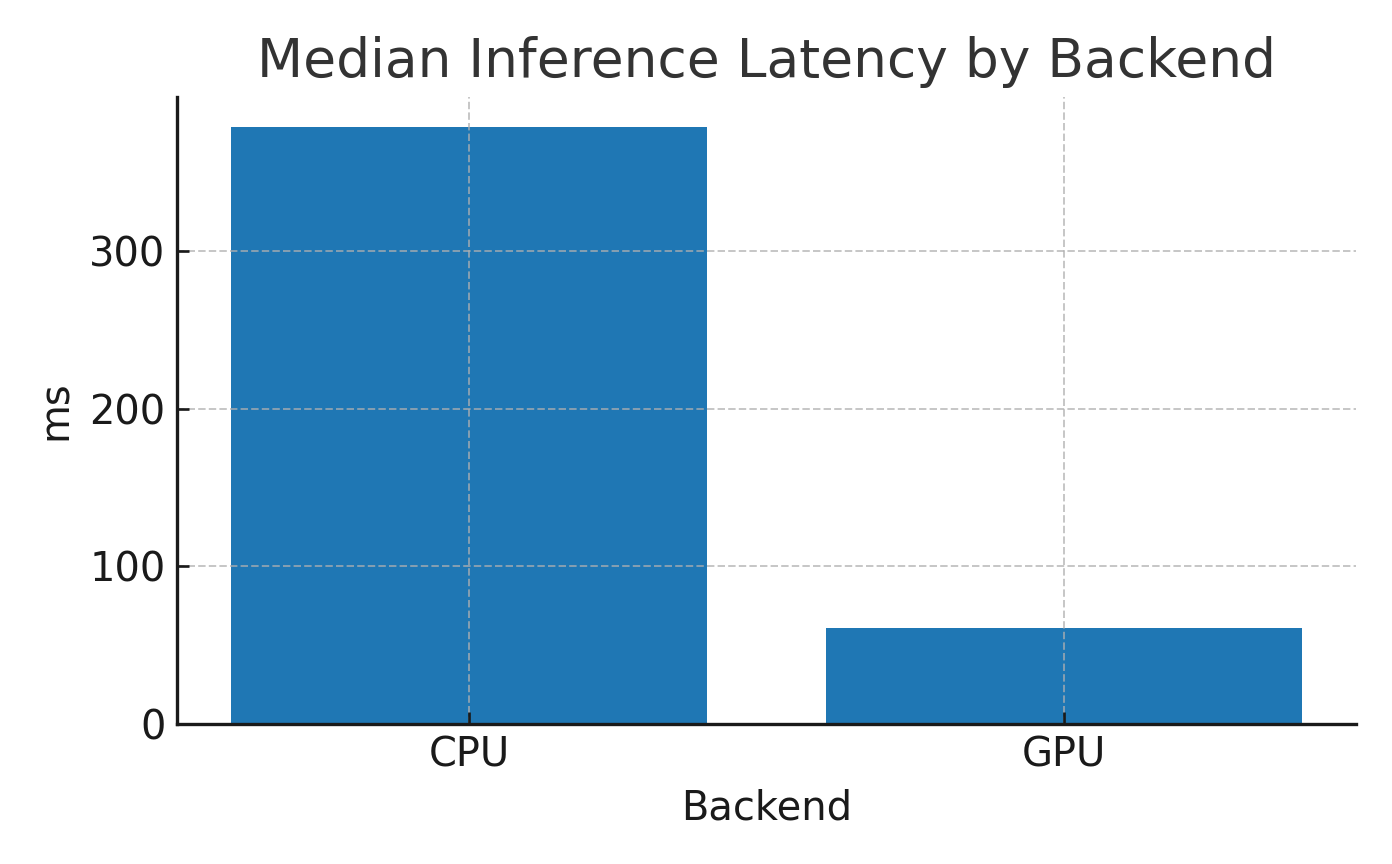}{Median inference latency by backend.}{fig:latency}{5.0cm}

\noindent\textit{Finding.} Compiled backends reduce median latency for several architectures, illustrating an accuracy--performance trade-off when compatibility gaps appear---underscoring the value of targeted stabilizers (deterministic adapters, fallbacks).

\subsection{Overall Summary}
\begin{table}[H]
\centering
\caption{Overall experiment summary.}
\label{tab:overall}
\begin{tabular}{@{}lcl@{}}
\toprule
\textbf{Metric} & \textbf{Value} & \textbf{Notes} \\
\midrule
Total checks & 672 & four $\mathrm{atol}$ settings \\
Passed & 484 & aggregate across models/backends \\
Pass rate & 72.0\% & overall \\
Distinct models & 19 & classification/detection/segmentation \\
Target backends & 2 & GPU (eager), compiled \\
Distinct $\mathrm{atol}$ & 4 & $\{10^{-6},10^{-5},10^{-4},10^{-3}\}$ \\
\bottomrule
\end{tabular}
\end{table}

% ===================== CASE STUDIES =====================
\section{Case Study: Detection Drift from Nondeterministic NMS}
\textbf{Symptom.} CPU vs.\ GPU/compiled runs exhibit inconsistent pre-NMS box ordering, leading to task-level discrepancies despite small tensor-level differences.\\
\textbf{Observation.} Agreement remains high up to the detection head; deviations emerge at the pre-NMS ordering stage, consistent with nondeterministic tie-breaking rather than upstream feature misalignment.\\
\textbf{Mitigation.} Deterministic sort over $(\text{score}, x_1, y_1)$ prior to NMS; alternatively, enforce FP32 for unstable kernels.\\
\textbf{Re-validation.} At $\mathrm{atol}=10^{-5}$, deterministic sorting restores task-level agreement without degrading latency benefits from compilation.

\PlaceBeforeAfterFigure{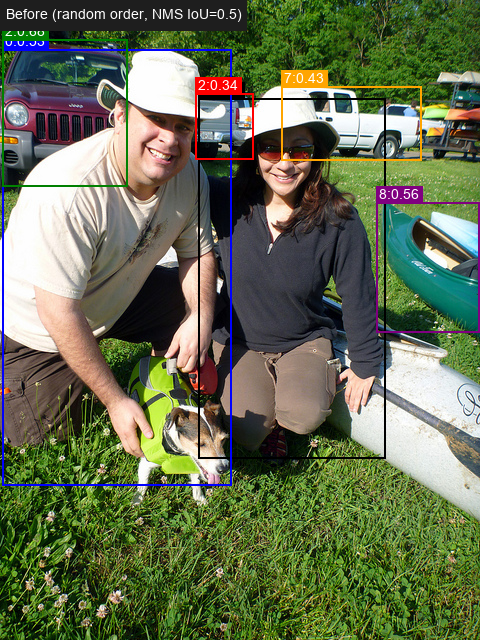}{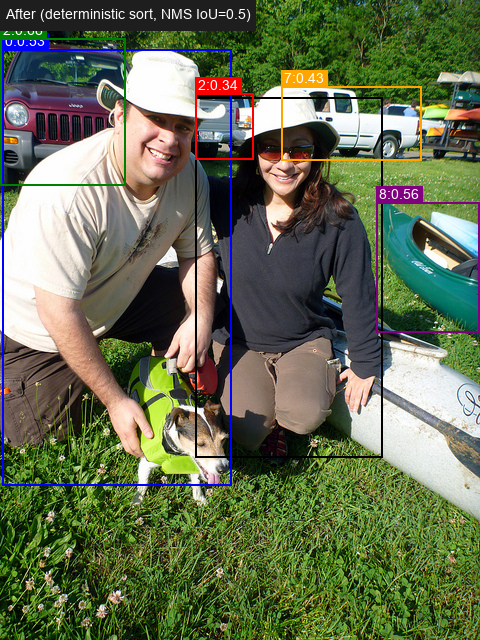}{Qualitative detection comparison on the same image: left shows inconsistent NMS outcomes across backends; right shows alignment after enforcing deterministic pre-NMS sorting.}{fig:case1}{5.2cm}

% ===================== DISCUSSION =====================
\section{Discussion}
\textbf{Threats to Validity.} Hardware/driver autotuning, precision modes (AMP vs.\ FP32), preprocessing mismatches (resize/interp), and batch-size effects may introduce residual nondeterminism. While we log environment fingerprints, remaining variance cannot be fully excluded.

\textbf{Lessons.}
\begin{itemize}[leftmargin=1.2em]
  \item \textbf{Tolerance calibration.} Very tight thresholds ($10^{-6}$) surface numerically small yet order-sensitive perturbations; thresholds should reflect end-task tolerance.
  \item \textbf{Deterministic adapters.} Sorting candidates before NMS removes order-induced divergence in detection with negligible overhead.
  \item \textbf{Operator fallbacks.} For problematic kernels, selective eager/FP32 fallbacks improve stability while preserving most performance gains.
\end{itemize}

\textbf{Future Work.} We plan a systematic activation-level survey across architectures and backends; broader model families including generative and multimodal; and additional runtimes (ONNX Runtime, TensorRT). We also aim to integrate the framework with reliable, efficient \emph{foundation models}, aligning with emerging research priorities in reproducible, cross-platform deployment.

% ===================== REPRO & CONCLUSION =====================
\section{Reproducibility and Artifacts}
We provide sanitized code, environment lockfiles, and scripts to regenerate JSONL logs and all tables/figures. An anonymized artifact is available for review and will be released publicly upon acceptance. The updated implementation is available at \url{https://github.com/william-zehua-li/cross-backend-model-checker}. All experiments can be reproduced with the provided configs and scripts.

\section{Conclusion}
We introduced a configuration-first framework for assessing cross-backend compatibility with a three-tier verifier that links tensor-level drift to task-level outcomes. Across $672$ checks spanning four tolerance settings, $72.0\%$ of runs pass; enforcing deterministic pre-NMS sorting restores detection-level agreement without forfeiting the latency benefits of compilation. \emph{To our knowledge, this is the first unified framework that systematically quantifies and mitigates cross-backend drift under a common configuration-first protocol.} We believe this advances dependable deployment of deep learning in safety-critical domains—such as medical imaging and autonomous systems—where cross-backend consistency is essential for reproducibility and assurance.

% ===================== BIBLIO (stub) =====================
\bibliographystyle{IEEEtran}

% ===================== APPENDIX =====================
\appendix

\section{Config Examples (Library/Repo)}
\begin{lstlisting}[style=pycode]
# Library example (classification)
from: library
model: resnet18
inputs: [assets/cat.jpg]
means: [0.485, 0.456, 0.406]
stds:  [0.229, 0.224, 0.225]
options: { compile: true, resize_multiple: 32 }
verification: { tol: { atol: 1e-5, rtol: 1e-5 } }
\end{lstlisting}

\begin{lstlisting}[style=pycode]
# Repo example (local clone + class path)
from: repo
repo: { path: ../third_party/yolo_clone, class: models.yolo.YoloNet }
params: { img_size: 640 }
inputs: [data/dog.jpg]
\end{lstlisting}

\section{sanitized\_main.py}
\begin{lstlisting}[style=pycode]
#!/usr/bin/env python3
import argparse, glob, os
from typing import Any, List, Tuple, Union
import torch
from PIL import Image
from sanitized_utils import (
    set_global_seed, load_yaml, resolve_path, load_image_tensor,
    adjust_to_multiple, tensors_allclose, to_cpu_like,
)
from sanitized_loaders import LibraryLoader, RepoLoader
TensorOrList = Union[torch.Tensor, List[torch.Tensor]]

def _normalize_output(y: Any) -> List[torch.Tensor]:
    if isinstance(y, torch.Tensor): return [y]
    if isinstance(y, (list, tuple)): return [t for t in y if isinstance(t, torch.Tensor)]
    if isinstance(y, dict): return [v for v in y.values() if isinstance(v, torch.Tensor)]
    return []

def run_once(cfg_path: str, device: str, use_compile: bool) -> Tuple[TensorOrList, TensorOrList]:
    cfg = load_yaml(cfg_path)
    source = cfg.get("from", "library")
    means = cfg.get("means", [0.485, 0.456, 0.406])
    stds  = cfg.get("stds",  [0.229, 0.224, 0.225])
    inputs = cfg.get("inputs", [])
    options = cfg.get("options", {})
    resize_multiple = options.get("resize_multiple", 32)
    if source == "library":
        model_name = cfg.get("model")
        if not model_name: raise ValueError("For 'library' source you must specify 'model'.")
        loader = LibraryLoader(model_name=model_name, weights=cfg.get("weights"))
        model = loader.build()
    elif source == "repo":
        repo = cfg.get("repo", {})
        repo_path = resolve_path(cfg_path, repo.get("path", ""))
        class_path = repo.get("class", "")
        params = cfg.get("params", {})
        loader = RepoLoader(repo_path=repo_path, class_path=class_path, params=params)
        model = loader.build()
    else:
        raise ValueError(f"Unknown source: {source}")
    model.eval()
    ref_device = torch.device("cpu")
    model_ref = model.to(ref_device)
    tgt_device = torch.device(device)
    model_tgt = model.to(tgt_device)
    if use_compile: model_tgt = torch.compile(model_tgt)
    ref_outputs: List[torch.Tensor] = []
    tgt_outputs: List[torch.Tensor] = []
    for rel in inputs:
        img_path = resolve_path(cfg_path, rel)
        img = Image.open(img_path).convert("RGB")
        x = load_image_tensor(img, means, stds).unsqueeze(0)
        x_ref = x.to(ref_device); x_tgt = x.to(tgt_device)
        if resize_multiple:
            x_ref = adjust_to_multiple(x_ref, resize_multiple)
            x_tgt = adjust_to_multiple(x_tgt, resize_multiple)
        with torch.no_grad():
            y_ref = model_ref(x_ref); y_tgt = model_tgt(x_tgt)
        ref_outputs.extend([to_cpu_like(t) for t in _normalize_output(y_ref)])
        tgt_outputs.extend([to_cpu_like(t) for t in _normalize_output(y_tgt)])
    return (ref_outputs if len(ref_outputs) != 1 else ref_outputs[0],
            tgt_outputs if len(tgt_outputs) != 1 else tgt_outputs[0])

def main():
    parser = argparse.ArgumentParser(description="Sanitized compatibility test runner")
    parser.add_argument("-d", "--device", required=True, help="Target device, e.g. cpu or cuda")
    parser.add_argument("-c", "--configs", default="configs/*.yaml", help="Glob for YAML configs")
    parser.add_argument("--compile", action="store_true", help="Use torch.compile for target run")
    parser.add_argument("--seed", type=int, default=5, help="Global RNG seed")
    args = parser.parse_args()
    set_global_seed(args.seed)
    cfg_files = sorted(glob.glob(args.configs))
    if not cfg_files:
        print(f"No configs matched: {args.configs}"); return
    total, passed, failed = 0, 0, 0
    for cfg in cfg_files:
        total += 1
        try:
            ref, tgt = run_once(cfg, args.device, args.compile)
            conf = load_yaml(cfg)
            tol = (((conf.get("verification") or {}).get("tol")) or {})
            atol = float(tol.get("atol", 1e-5)); rtol = float(tol.get("rtol", 1e-5))
            ok = tensors_allclose(ref, tgt, atol=atol, rtol=rtol)
            status = "PASS" if ok else "FAIL"
            if ok: passed += 1
            else: failed += 1
            print(f"[{status}] {os.path.basename(cfg)} (atol={atol}, rtol={rtol})")
        except Exception as e:
            failed += 1
            print(f"[ERROR] {os.path.basename(cfg)} -> {e}")
    print("\n=== Summary ==="); print(f"Total: {total}  Passed: {passed}  Failed: {failed}")

if __name__ == "__main__": main()
\end{lstlisting}

\section{sanitized\_loaders.py}
\begin{lstlisting}[style=pycode]
# Sanitized loaders: only public sources; no proprietary modules.
import importlib, os
from typing import Any, Dict, Optional
import torch, torch.nn as nn

class LibraryLoader:
    def __init__(self, model_name: str, weights: Optional[str] = None, params: Optional[Dict[str, Any]] = None):
        self.model_name = model_name; self.weights = weights; self.params = params or {}
    def build(self) -> nn.Module:
        from torchvision import models
        if not hasattr(models, self.model_name):
            raise ValueError(f"Unknown library model: {self.model_name}")
        ctor = getattr(models, self.model_name); model = ctor(**self.params)
        if self.weights and os.path.exists(self.weights):
            state = torch.load(self.weights, map_location="cpu")
            if isinstance(state, dict) and "state_dict" in state: state = state["state_dict"]
            model.load_state_dict(state, strict=False)
        return model

class RepoLoader:
    def __init__(self, repo_path: str, class_path: str, params: Optional[Dict[str, Any]] = None):
        if not repo_path or not os.path.isdir(repo_path): raise FileNotFoundError(f"repo_path not found: {repo_path}")
        if "." not in class_path: raise ValueError("class_path must be dotted, e.g. 'pkg.subpkg.Class'")
        self.repo_path = os.path.abspath(repo_path); self.class_path = class_path; self.params = params or {}
    def build(self) -> nn.Module:
        import sys
        sys.path.insert(0, self.repo_path)
        try:
            module_path, cls_name = self.class_path.rsplit(".", 1)
            module = importlib.import_module(module_path); cls = getattr(module, cls_name)
            model = cls(**self.params); return model
        finally:
            if self.repo_path in sys.path: sys.path.remove(self.repo_path)
\end{lstlisting}

\section{sanitized\_utils.py}
\begin{lstlisting}[style=pycode]
# Utilities for the sanitized runner.
import os, random
from typing import List, Sequence, Union
import numpy as np, torch
from PIL import Image
import yaml
from torchvision import transforms

def set_global_seed(seed: int) -> None:
    os.environ["PYTHONHASHSEED"] = str(seed); random.seed(seed); np.random.seed(seed)
    torch.manual_seed(seed); torch.cuda.manual_seed_all(seed)
    torch.backends.cudnn.deterministic = True; torch.backends.cudnn.benchmark = False

def load_yaml(path: str) -> dict:
    with open(path, "r", encoding="utf-8") as f: return yaml.safe_load(f) or {}

def resolve_path(base_cfg: str, rel: str) -> str:
    if os.path.isabs(rel): return rel
    base_dir = os.path.dirname(os.path.abspath(base_cfg)); return os.path.normpath(os.path.join(base_dir, rel))

def load_image_tensor(img: Image.Image, means: Sequence[float], stds: Sequence[float]) -> torch.Tensor:
    pre = transforms.Compose([transforms.Resize(256), transforms.CenterCrop(224), transforms.ToTensor(),
                              transforms.Normalize(mean=list(means), std=list(stds)),])
    return pre(img)

def adjust_to_multiple(x: torch.Tensor, m: int) -> torch.Tensor:
    if x.dim() != 4: return x
    _, _, h, w = x.shape; nh = max(m, (h // m) * m); nw = max(m, (w // m) * m)
    if nh == h and nw == w: return x
    return torch.nn.functional.interpolate(x, size=(nh, nw), mode="bilinear", align_corners=False)

def to_cpu_like(t: torch.Tensor) -> torch.Tensor: return t.detach().to("cpu")

def _allclose(a: torch.Tensor, b: torch.Tensor, atol: float, rtol: float) -> bool:
    try: torch.testing.assert_close(a, b, atol=atol, rtol=rtol); return True
    except AssertionError: return False

def tensors_allclose(a: Union[torch.Tensor, List[torch.Tensor]], b: Union[torch.Tensor, List[torch.Tensor]], atol: float = 1e-5, rtol: float = 1e-5) -> bool:
    if isinstance(a, torch.Tensor) and isinstance(b, torch.Tensor): return _allclose(a, b, atol, rtol)
    if isinstance(a, list) and isinstance(b, list):
        if len(a) != len(b): return False
        for ta, tb in zip(a, b):
            if not _allclose(ta, tb, atol, rtol): return False
        return True
    return False
\end{lstlisting}

\section{result\_logger.py (Optional)}
\begin{lstlisting}[style=pycode]
# Minimal JSONL logger to unify outputs for paper tables/figures.
import json, time, os, sys
from typing import Any, Dict

def log_jsonl(record: Dict[str, Any], path: str) -> None:
    rec = dict(record)
    rec["timestamp"] = time.strftime("%Y-%m-%dT%H:%M:%S")
    os.makedirs(os.path.dirname(path), exist_ok=True)
    with open(path, "a", encoding="utf-8") as f:
        f.write(json.dumps(rec, ensure_ascii=False) + "\n")

def log_run(config_path: str, model: str, backend_pair: str,
            atol: float, rtol: float, status: str, stats: Dict[str, Any], out_path: str):
    record = {
        "config": config_path, "model": model, "backend_pair": backend_pair,
        "atol": atol, "rtol": rtol, "status": status, **(stats or {})
    }
    log_jsonl(record, out_path)
\end{lstlisting}

\end{document}